\documentclass[sigconf]{acmart}

\usepackage{amsmath}
\usepackage{algorithm,algpseudocode}
\DeclareMathOperator*{\argmax}{argmax}

\AtBeginDocument{%
  \providecommand\BibTeX{{%
    \normalfont B\kern-0.5em{\scshape i\kern-0.25em b}\kern-0.8em\TeX}}}

\copyrightyear{2020}
\acmYear{2020}
\setcopyright{acmcopyright}\acmConference[KDD '20]{Proceedings of the 26th ACM SIGKDD Conference on Knowledge Discovery and Data Mining}{August 23--27, 2020}{Virtual Event, CA, USA}
\acmBooktitle{Proceedings of the 26th ACM SIGKDD Conference on Knowledge Discovery and Data Mining (KDD '20), August 23--27, 2020, Virtual Event, CA, USA}
\acmPrice{15.00}
\acmDOI{10.1145/3394486.3403085}
\acmISBN{978-1-4503-7998-4/20/08}

\begin{document}
	\fancyhead{}
\title[XGNN: Towards Model-Level Explanations of Graph Neural Networks]{XGNN: Towards Model-Level Explanations of Graph \\Neural Networks}


\author{Hao Yuan}
\email{hao.yuan@tamu.edu}
\affiliation{%
  \institution{Texas A\&M University}
  \city{College Station}
  \state{Texas}
  \country{United States}
  \postcode{77840}
}

\author{Jiliang Tang}
\email{tangjili@msu.edu}
\affiliation{%
	\institution{Michigan State University}
	\city{East Lansing}
	\state{Michigan}
	\country{United States}
	\postcode{48824}
}

\author{Xia Hu}
\email{hu@cse.tamu.edu}
\affiliation{%
	\institution{Texas A\&M University}
	\city{College Station}
	\state{Texas}
	\country{United States}
	\postcode{77840}
}

\author{Shuiwang ji}
\email{sji@tamu.edu}
\affiliation{%
	\institution{Texas A\&M University}
	\city{College Station}
	\state{Texas}
	\country{United States}
	\postcode{77840}
}

\renewcommand{\shortauthors}{Trovato and Tobin, et al.}

\begin{abstract}
	Graphs neural networks (GNNs) learn node features by aggregating and combining neighbor information,
	which have achieved promising performance on many graph tasks. However, GNNs are mostly treated as black-boxes and lack human
	intelligible explanations. Thus, they cannot be fully trusted and
	used in certain application domains if GNN models cannot be
	explained. In this work, we propose a novel approach, known as XGNN,
	to interpret GNNs at the model-level. Our approach can provide
	high-level insights and generic understanding of how GNNs work. In
	particular, we propose to explain GNNs by training a graph generator
	so that the generated graph patterns maximize a certain prediction
	of the model. We formulate the graph generation as a reinforcement
	learning task, where for each step, the graph generator predicts how
	to add an edge into the current graph. The graph generator is
	trained via a policy gradient method based on information from the
	trained GNNs. In addition, we incorporate several graph rules to
	encourage the generated graphs to be valid. Experimental results 
	on both synthetic and real-world datasets show that our proposed methods
	help understand and verify the trained GNNs. Furthermore, our
	experimental results indicate that the generated graphs can provide
	guidance on how to improve the trained GNNs.
\end{abstract}




\maketitle

\section{Introduction}
Graph Neural Networks (GNNs) have shown their effectiveness and
obtained the state-of-the-art performance on different graph tasks, such
as node classification~\cite{gao2018graph, velivckovic2017graph},
graph  classification~\cite{xu2018powerful, zhang2018end}, and link
prediction~\cite{zhang2018link}. In addition, extensive efforts have been made towards 
different graph operations, such as graph convolution~\cite{kipf2016semi, gilmer2017neural, hamilton2017inductive}, graph pooling~\cite{Yuan2020StructPool,lee2019self}, and graph attention~\cite{velivckovic2017graph, thekumparampil2018attention, gao2019graph}. 
Since graph
data widely exist in different real-world applications, such as
social networks, chemistry, and biology, GNNs are becoming
increasingly important and useful. Despite their great performance,
GNNs share the same drawback as other deep learning models; that is,
they are usually treated as black-boxes and lack human-intelligible
explanations. 
Without understanding and verifying the inner working
mechanisms, GNNs cannot be fully trusted, which prevents their use
in critical applications pertaining to fairness, privacy, and
safety~\cite{doshi2017towards, ying2019gnn}. 
For example, we can train a GNN model to predict the effects of drugs where we treat each drug as a molecular graph. 
Without exploring the working mechanisms, we do not know what chemical groups in a molecular graph lead to the predictions.    
Then we cannot verify whether the rules of the GNN model are consistent with real-world chemical rules, and hence we cannot fully trust the GNN model.
This raises the need of
developing interpretation techniques for GNNs.

Recently, several interpretations techniques have been proposed to explain deep learning models
on image and text data.
Depending on what kind of interpretations are provided,
existing techniques can be categorized into
example-level~\cite{simonyan2013deep,smilkov2017smoothgrad,
	yuan2019interpreting,selvaraju2017grad,zhou2016learning,zeiler2014visualizing,fong2017interpretable,dabkowski2017real}
or model-level~\cite{erhan2009visualizing,nguyen2017plug,nguyen2015deep}
methods.  
Example-level interpretations explain the prediction for a
given input example, by determining important features in the input
or the decision procedure for this input through the model. 
Common
techniques in this category include gradient-based
methods~\cite{simonyan2013deep,smilkov2017smoothgrad,yuan2019interpreting},
visualizations of intermediate feature
maps~\cite{selvaraju2017grad,zhou2016learning}, and occlusion-based
methods~\cite{zeiler2014visualizing,fong2017interpretable,dabkowski2017real}.
Instead of providing input-dependent
explanations, model-level interpretations aim to explain the general
behavior of the model by investigating what input patterns can lead
to a certain prediction, without respect to any specific input
example. 
Input optimization~\cite{erhan2009visualizing,nguyen2017plug,nguyen2015deep,olah2017feature}
is the most popular model-level interpretation method. 
These two categories of interpretation methods aim at explaining deep models in different views. 
Since the ultimate goal of interpretations is to verify and understand deep models, 
we need to manually check the interpretation results and conclude if the deep models work in our expected way.
For example-level 
methods, we may need to explore the explanations for a large number of examples before we can trust the models. 
However, it is time-consuming and requires extensive expert efforts. For model-level methods, the explanations
are more general and high-level, and hence need less human supervision. However, the explanations of model-level methods
are less precise compared with example-level interpretations. Overall, both  model-level and example-level methods are 
important for interpreting and understanding deep models.

Interpreting deep learning models on graph data become increasingly important but is still less explored. To the best of our knowledge, 
there is no existing study on interpreting GNNs at the model-level. The existing study~\cite{ying2019gnn, baldassarre2019explainability} only provides example-level explanations for graph models.  As a radical departure from existing work, we propose a novel interpretation technique, known as XGNN, for explaining deep graph models at the model-level. We propose to investigate what graph patterns can maximize a certain prediction. Specifically, we propose to train a graph generator such that the generated graph patterns can be used to explain deep graph models. We formulate it as a reinforcement learning problem that at each step, the graph generator predicts how to add an edge to a given graph and form a new graph. Then the generator is trained based on the feedback from the trained graph models using policy gradient~\cite{sutton2000policy}.
We also incorporate several graph rules to encourage the generated graphs to be valid. Note that the graph generation part in our XGNN framework can be generalized to any suitable graph generation method, determined by the dataset at hand and the GNNs to be interpreted.
Finally, we trained GNN models on both real-world and synthetic datasets which can yield good performance. Then we employ our proposed XGNN to explain these trained models. Experimental results show that our proposed XGNN can find the desired graph patterns and explains these models. With our generated graph patterns, we can verify, understand, and even improve the trained GNN models.   

\section{Related Work}
\subsection{Graph Neural Networks}
Graphs are wildly employed to represent data in different real-world domains and graph neural networks have shown promising performance on these data. Different from image and text data, a graph is represented by a feature matrix and an adjacency matrix. Formally, a graph $G$ with $n$ nodes is represented by its feature matrix $X \in \mathbb{R}^{n\times d}$ and its adjacency matrix $A \in \{0,1\}^{n\times n}$. Note that we assume each node has a $d$-dimension vector to represent its features. Graph neural networks learn node features based on these matrices. Even though there are several variants of GNNs, such as graph convolution networks (GCNs)~\cite{kipf2016semi}, graph attention networks (GATs)~\cite{velivckovic2017graph}, and graph isomorphism networks (GINs)~\cite{xu2018powerful}, they share a similar feature learning strategy. For each node, GNNs update its node features by aggregating the features from its neighbors and combining them with its own features. We take GCNs as an example to illustrate the neighborhood information aggregation scheme. The operation of GCNs is defined as 
\begin{equation}\label{eq:1}
X_{i+1} = f(D^{-\frac{1}{2}}\hat{A}D^{-\frac{1}{2}}X_{i}W_{i}),
\end{equation}
where $X_{i} \in \mathbb{R}^{n\times d_{i}}$ and $X_{i+1} \in \mathbb{R}^{n\times d_{i+1}}$ are the input and output feature matrices of the $i^{th}$ graph convolution layer. In addition, $\hat{A}=A+I$ is used to add self-loops to the adjacency matrix, $D$ denotes the diagonal node degree matrix to normalize $\hat{A}$. The matrix 
$W_{i}\in \mathbb{R}^{d_{i}\times d_{i+1}}$ is a trainable matrix for layer $i$ and is used to perform linear feature transformation 
and $f(\cdot)$ denotes a non-linear activation function. By stacking $j$ graph convolution layers, the $j$-hop neighborhood information can be aggregated. Due to its superior performance, we incorporate the graph convolution in Equation~(\ref{eq:1}) as our graph neural network operator.

\subsection{Model-level Interpretations}\label{ml-ch}
Next, we briefly discuss popular model-level interpretation techniques for deep learning models
on image data, known as input optimization methods~\cite{erhan2009visualizing,nguyen2017plug,nguyen2015deep,olah2017feature}. These methods generally generate optimized input that can maximize a certain behavior of deep models. They randomly initialize the input and iteratively update the input towards an objective, such as maximizing a class score. Then such optimized input can be regarded as the explanations for the target behavior. Such a procedure is known as optimization and is similar to training deep neural networks. The main difference is that in such input optimization techniques, all network parameters are fixed while the input is treated as trainable variables. 
While such methods can provide meaningful model-level explanations
for deep models on images, they cannot be directly applied to
interpret GNNs due to three challenges. First, the structural information of a graph is represented by a
discrete adjacency matrix, which cannot be directly optimized via
back-propagation. Second, for images, the optimized input is an
abstract image and the visualization shows high-level patterns and meanings. In
the case of graphs, the abstract graph is not meaningful and hard to
visualize. 
Third, the obtained graphs may not be valid for chemical
or biological rules since non-differentiable graph rules cannot
be directly incorporated into optimization. For example, 
the node degree of an atom should not exceed its maximum chemical valency.

\subsection{Graph Model Interpretations}
To the best of our knowledge, there are only a few existing studies focusing on the interpretability of deep graph models~\cite{ying2019gnn, baldassarre2019explainability}.
The recent GNN interpretation tool GNN Explainer~\cite{ying2019gnn} proposes to explain deep graph models at the example-level by learning soft masks. For a given example, it applies soft masks to graph edges and node features and updates the masks such that the prediction remains the same as the original one. 
Then some graph edges and node features are selected by thresholding the masks, and they are treated as important edges and features for making the
prediction for the given example. 
\begin{figure*}[!ht]
	\centering
	\includegraphics[width=0.9\textwidth]{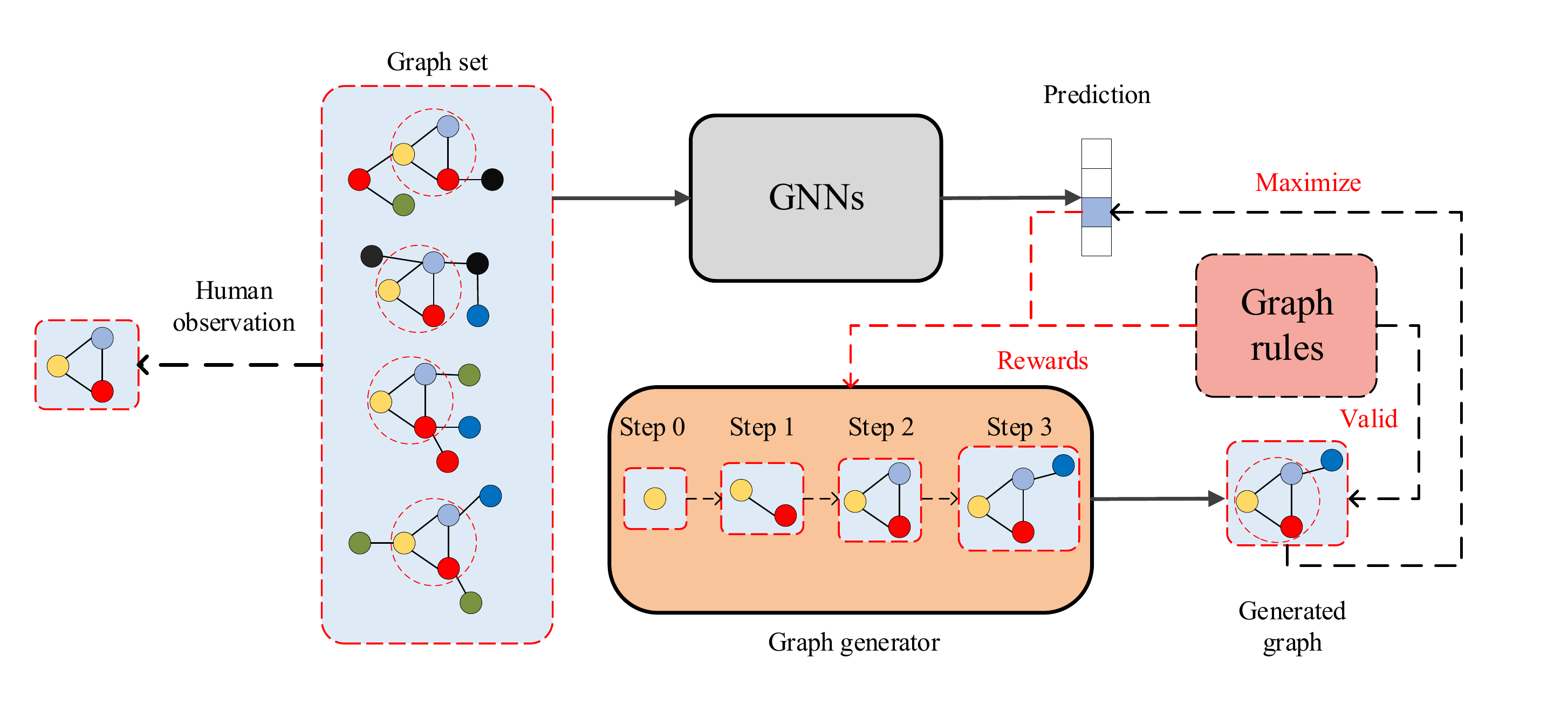}
	\caption{Illustrations of our proposed XGNN for graph
		interpretation via graph generation. The GNNs represent a trained
		graph classification model that we try to explain. All graph
		examples in the graph set are  classified to the third class. 
		The	left part shows that we can manually conclude the key graph patterns
		for the third class but it is challenging. The right part shows that we propose to train a graph
		generator to generate graphs that can maximize the class score and
		be valid according to graph rules.}
	\label{fig:pip}
\end{figure*}
The other work~\cite{baldassarre2019explainability} also focuses on the example-level interpretations of deep graph models. It applies several well-known image interpretation methods to graph models, such as sensitivity analysis (SA)~\cite{gevrey2003review}, guided backpropagation (GBP)~\cite{springenberg2014striving}, and layer-wise relevance propagation (LRP)~\cite{bach2015pixel}. The  SA and GBP methods are based on the gradients while the LRP method computes the saliency maps by decomposing the output prediction into a combination of its inputs. 
In addition, both of these studies generate input-dependent explanations for individual examples. To verify and understand a deep model, 
humans need to check explanations for all examples, which is time-consuming or even not feasible.

While input-dependent explanations are important for understanding deep models, model-level interpretations should not be ignored. However, none of the existing work investigates the model-level interpretations of deep graph models. In this work, we argue that model-level interpretations can provide higher-level insights and a more general understanding in how a deep learning model works. 
Therefore, we aim at providing model-level interpretations for GNNs. We propose a novel method, known as XGNN,  to explain GNNs by graph generation such that the generated graphs can maximize a certain behavior.   
\section{XGNN: Explainable Graph Neural Networks}
\subsection{Model-Level GNN Interpretation}\label{model-lv}
Intuitively, given a trained GNN  model, the 
model-level interpretations for it should explain
what graph patterns or sub-graph patterns lead to a certain
prediction. For example, one possible type of patterns is known as network motifs that
represent simple building blocks of complex networks (graphs), which
widely exist in graphs from biochemistry, neurobiology, ecology, and
engineering~\cite{milo2002network,alon2006introduction,alon2007network,shen2002network}.
Different motif sets can be found in graphs with different
functions~\cite{milo2002network, alon2006introduction}, which means different motifs may directly relate to the functions of graphs. 
However, it is still unknown whether GNNs make predictions based on such motifs or other graph information. By identifying the relationships between graph patterns and the predictions of GNNs, we can better understand the models and verify whether a model works as expected. Therefore, we propose our XGNN, which explains GNNs using such graph patterns. 
Specifically, in this work, we investigate the model-level interpretations of GNNs for graph classification tasks and the graph patterns are obtained by graph generations.

Formally, let $f(\cdot)$ denote a trained GNN classification
model, and $y \in \{c_1, \cdots, c_{\ell}\}$ denote the classification
prediction. Given $f(\cdot)$ and a chosen class $c_i$, $i\in \{1, \cdots,
\ell\}$, our goal is to investigate what input graph patterns maximize the
predicted probability for this class. The obtained patterns can be
treated as model-level interpretations with respect to $c_i$.
Formally, the  task can be defined as
\begin{equation}\label{form1}
G^*= \argmax_{G} P(f(G)=c_i),
\end{equation}
where $G^*$ is the optimized input graph we need.  A popular way to obtain such optimized input for interpreting image and text models is known as input optimization~\cite{yuan2019interpreting,erhan2009visualizing,nguyen2017plug,nguyen2015deep,olah2017feature}. However, as discussed in Section~\ref{ml-ch}, such optimization method cannot be applied to interpret graph models because of the special representations of graph data. Instead, we propose to obtain the optimized graph $G^*$ via graph generation.   
The general illustration of our proposed method is shown in Figure~\ref{fig:pip}. Given a pre-trained graph classification model, we interpret it by providing explanations for its third class. We may manually conclude the graph patterns from the graph dataset. By evaluating all graph examples in the dataset, we can obtain the graphs that are predicted to be the third class. Then we can manually check what are the common graph patterns among these graphs. For example, the left part of Figure~\ref{fig:pip} shows that a set of four graphs are
classified into the third class. Based on human observations, we know
that the important graph pattern leading to the prediction is the
triangle pattern consisting of a red node, a yellow node, and a blue
node.
However, such manual analysis is time-consuming and not applicable for large-scale and complex graph datasets. As shown in the right part, we propose to train a graph generator to generate graph patterns that can maximize the prediction score of the third class. In addition, we incorporate graph rules, such as the chemical valency check, to encourage valid and human-intelligible explanations. Finally, we can analyze the generated graphs to obtain model-level explanations for the third class. Compared with directly manual analysis on the original dataset, our proposed method generates small-scale and less complex graphs, which can significantly reduce the cost for further manual analysis.

\subsection{Interpreting GNNs via Graph Generation}\label{sec:rl}
Recent advances in graph generation lead to many successful graph generation models,
such as GraphGAN~\cite{wang2018graphgan},
ORGAN~\cite{guimaraes2017objective}, Junction Tree
VAE~\cite{jin2018junction}, DGMG~\cite{li2018learning}, and Graph Convolutional Policy Network
(GCPN)~\cite{you2018graph}. 
Inspired by these methods, we propose to train a graph generator which generates $G^*$ step by step. 
For each step, the graph generator generates a new graph based on the current graph.  
Formally, we define the partially generated graph at step $t$ as $G_t$, which contains $n_t$ nodes. 
It is represented as a feature matrix $X_t \in \mathbb{R}^{n_t\times d}$ and an adjacency matrix $A_t \in \{0,1\}^{n_t\times n_t}$, 
assuming each node has a $d$-dimensional feature vector. Then we
define a $\theta$-parameterized graph generator as $g_\theta(\cdot)$, which takes $G_{t}$ as input, and outputs a new graph $G_{t+1}$ that
\begin{equation}\label{form2}
X_{t+1}, A_{t+1}=  g_\theta(X_t, A_t).
\end{equation}

Then the generator is trained with the guidance from the pre-trained GNNs $f(\cdot)$. Since generating the new graph $G_{t+1}$ from $G_{t}$ is non-differentiable, we formulate the generation procedure as a reinforcement learning problem. 
Specifically, assuming there are $k$ types of nodes in the dataset, we define a candidate set $C=\{s_1, s_2, \cdots, s_k\}$ denoting these possible node types. 
For example,
in a chemical molecular dataset, the candidate set can be $C=\{Carbon,
Nitrogen, \cdots, Oxygen,  Fluorine \}$. In a social network dataset where nodes are not labeled, the candidate set only contains a single node type. Then at each step $t$, based on the partially generated graph $G_t$, the generator $g(\cdot)$ generates $G_{t+1}$ by predicting how to add an edge to the current graph $G_t$. Note that the generator may add an edge between two nodes in the current graph $G_t$ or add a node from the candidate set $C$ to the current graph $G_t$ and connect it with an existing node in  $G_t$. Formally, we formulate it as a reinforcement learning problem, which consists of four elements: state, action, policy, and reward.

\textbf{State}: The state of the reinforcement learning environment at step $t$ is the partially generated graph $G_t$. The initial graph at the first step can be either a random node from the candidate set $C$ or manually designed based on prior domain knowledge. For example, for the dataset describing organic molecules, we can set the initial graph as a single node labeled with carbon atom since any organic compound contains carbon generally~\cite{seager2013chemistry}.

\textbf{Action}: The action at step $t$, denoted as $a_t$, is to generate the new graph $G_{t+1}$ based on the current graph $G_t$. Specifically, given the current state $G_t$, the action $a_t$ is to add an edge to $G_t$ by determining the starting node and the ending node of the edge. Note that the starting node $a_{t,start}$ can be 
any node from the current graph $G_t$ while the ending node $a_{t,end}$ is selected from the union of the current graph $G_t$ and the candidate set $C$ excluding the selected starting node $a_{t,start}$, denoted as $(G_t \bigcup C)\setminus a_{t,start}$. Note that with the predefined maximum action step and maximum node number, we can control the termination of graph generation. 

\textbf{Policy}: We employ graph neural networks to serve as the policy. The policy determines the action $a_t$ based on the state $G_t$. Specifically, the policy is the graph generator $g_\theta(\cdot)$, which takes $G_t$ and $C$ as the input and outputs the probabilities of possible actions. With the reward function, the generator $g_\theta(\cdot)$ can be trained via policy gradient~\cite{sutton2000policy}.

\textbf{Reward}: The reward for step $t$, denoted as $R_t$, is employed to evaluate the action at step $t$, which consists of two parts. The first part is the guidance from the trained GNNs $f(\cdot)$, which encourages the generated graph to maximize the class score of class $c_i$. By feeding the generated graphs to $f(\cdot)$, we can obtain the predicted probabilities for class $c_i$ and use them as the feedback to update $g_\theta(\cdot)$. 
The second part encourages the generated graphs to be valid in terms of certain graph rules. For example, for social network datasets, it is may not allowed to add multiple edges between two nodes. In addition, for chemical molecular datasets, the degree of an atom cannot exceed its chemical valency. 
Note that for each step, we include both intermediate rewards and overall rewards to evaluate the action.  

While we formulate the graph generation as a reinforcement learning
problem, it is noteworthy that our proposed XGNN is a novel and
general framework for interpreting GNNs at the model-level. 
The graph generation part
in this framework can be generalized to any suitable graph
generation method, determined by the dataset at hand and the GNNs to be
interpreted.
\begin{figure*}[!ht]
	\centering
	\includegraphics[width=0.9\textwidth]{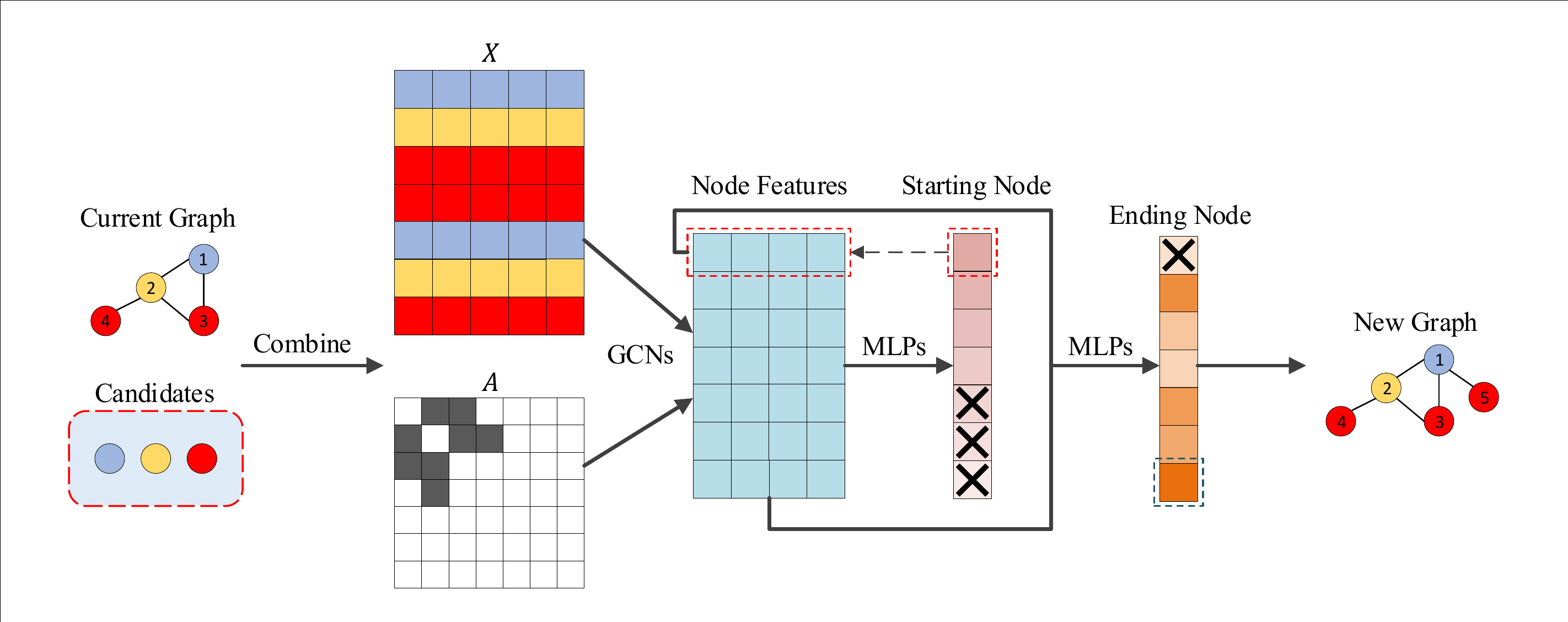}
	\caption{An Illustration of our graph generator for processing a single step. 
		Different colors denote different types of node. Given a graph with $4$ nodes and a candidate set with 3 nodes, 
		we first combine them together to obtain the feature matrix and the adjacency matrix. Then we employ several GCN layers to aggregate and learn node features. 
		Next, the first MLPs predict a probability distribution from which we sample the starting node. Finally, the second MLPs predict the ending node conditioned on the starting node. Note that the black crosses indicates masking out nodes. 
	}
	\label{fig:generator}
\end{figure*}
\subsection{Graph Generator}\label{sec:generator}

For step $t$, the graph generator $g_\theta(\cdot)$ incorporates the partially generated graph $G_t$ and the candidate set $C$ to predict the probabilities of different actions, denoted as $p_{t,start}$ and $p_{t,end}$.
Assume there are $n_t$ nodes in $G_t$ and $k$ nodes in $C$, then both $p_{t,start}$ and $p_{t,end}$ are with $n_t+k$ dimensionality.  
Then the action $a_t=(a_{t,start}, a_{t,end})$ is sampled from the probabilities $p_t=(p_{t,start}, p_{t,end})$. 
Next, we can obtain the new graph $G_{t+1}$ based on the action $a_t$. 
Specifically, in our generator, we first employ several graph convolutional layers to aggregate neighborhood information and learn node features. 
Mathematically, it can be written as
\begin{equation}\label{eqn1_3}
\widehat{X} = \mbox{GCNs}(G_t, C),
\end{equation}
where $\widehat{X}$ denotes the learnt node features. Note that the graph $G_t$ and the candidate set $C$ are combined as the input of GCNs. 
We merge all nodes in $C$ to $G_t$ without adding any edge and then obtain the new node feature matrix and adjacency matrix. 
Then Multilayer Perceptrons (MLPs) are used to predict the probabilities of the starting node, $p_{t,start}$ and the action $a_{t,start}$ is sampled from this probabilty distribution. Mathematically, it can be written as
\begin{eqnarray}\label{eqn1_4}
p_{t,start} &= &\mbox{Softmax}(\mbox{MLPs}(\widehat{X})),\\
a_{t,start} &\sim& p_{t,start}\cdot m_{t,start},
\end{eqnarray} 
where $\cdot$ means element-wise product and $m_{t,start}$ is  to mask out all candidate nodes since the starting node can be only selected from the current graph $G_t$.
Let $\widehat{x}_{start}$ denote the features of the node selected  by the start action $a_{t,start}$. Then conditioned on 
the selected node, we employ the second MLPs to compute the probability distribution of the ending node $p_{t,end}$ from which we sample the ending node action $a_{t,end}$. Note that since the starting node and the ending node cannot be the same, we apply a mask $m_{t,end}$ to mask out the node selected by $a_{t,start}$. 
Mathematically, it can be written as
\begin{eqnarray}\label{eqn1_5}
p_{t,end} &= &\mbox{Softmax}(\mbox{MLPs}([\widehat{X},\widehat{x}_{start}])),\\
a_{t,end} &\sim& p_{t,end}\cdot m_{t,end}\label{eqn1_6},
\end{eqnarray} 
where $[\cdot,\cdot]$ denotes broadcasting and concatenation. In addition, $m_{t,end}$ is the mask consisting of all 1s except the position indicating $a_{t,start}$. 
Note that the same graph generator $g_\theta(\cdot)$ is shared by different time steps, and our generator is capable to incorporate graphs with variable sizes.  

We illustrate our graph generator in Figure~\ref{fig:generator} where we show the graph generation procedure for one step. The current graph $G_t$ consists of 4 nodes and the candidate set has 3 available nodes. They are combined together to serve as the input of the graph generator. The embeddings of candidate nodes are concatenated to the feature matrix of $G_t$ while the adjacency matrix of $G_t$ is expanded accordingly. Then multiple graph convolutional layers are employed to learn features for all nodes. With the first MLPs, we obtain the probabilities of selecting different nodes as the starting node, and from which we sample the node 1 as the starting node. 
Then based on the features of node 1 and all node features, the second MLPs predict the ending node. We sample from the probabilities and select the node 7 as the ending node, which corresponds to the red node in the candidate set. Finally, a new graph is obtained by including a red node and connecting it with node 1. 

\subsection{Training the Graph Generator}\label{sec:reward}
The graph generator is trained to generate specific graphs that can maximize the class score of class $c_i$ and be valid to graph rules. Since such guidance is not differentiable, we employ policy gradient~\cite{sutton2000policy} to train the generator. According to~\cite{lei2016rationalizing, yu2017seqgan}, the loss function for the action $a_t$ at step $t$ can be mathematically written as
\begin{equation}\label{eq:ce}
\mathcal{L}_{g} = - R_t(\mathcal{L}_{CE}(p_{t,start}, a_{t,start})+ \mathcal{L}_{CE}(p_{t,end},a_{t,end})),
\end{equation}
where $\mathcal{L}_{CE}(\cdot,\cdot)$ denotes the cross entropy loss and $R_t$ means the reward function for step $t$. Intuitively, the reward $R_t$ indicates whether $a_t$ has a large chance to generate graph with high class score of class $c_i$ and being valid. Hence, the reward $R_t$ consists of two parts. The first part $R_{t,f}$ is the feedback from the trained model $f(\cdot)$ and the second part $R_{t,r}$ is from the graph rules. Specifically, for step $t$, the reward $R_{t,f}$  contains both an intermediate reward and a final graph reward for graph $G_{t+1}$ that 
\begin{equation}\label{eq:r1}
R_{t,f} = R_{t,f} (G_{t+1})+ \lambda_1 \frac{\sum_{i=1}^{m}R_{t,f}(\mbox{Rollout}({G_{t+1}}))}{m},
\end{equation}
where $\lambda_1$ is a hyper-parameter, and the first term is the intermediate reward which
can be obtained by feeding $G_{t+1}$ to the trained GNNs $f(\cdot)$ and checking the predicted probability for class $c_i$. Mathematically, it can be computed as
\begin{equation}\label{eq:r2}
R_{t,f} (G_{t+1}) = p(f(G_{t+1})=c_i) - 1/\ell,
\end{equation}
where $\ell$ denotes the number of possible classes for $f(\cdot)$. In addition, the second term in Equation (\ref{eq:r1}) is the final graph reward for $G_{t+1}$ which can be  obtained by performing Rollout~\cite{yu2017seqgan} $m$ times on the intermediate graph $G_{t+1}$. Each time, a final graph is generated based on $G_{t+1}$ until termination and then evaluated by $f(\cdot)$ using Equation (\ref{eq:r2}). Then the evaluations for $m$ final graphs are averaged to serve as the final graph reward.   
Overall, $R_{t,f}$ is positive when the obtained graph tends to yield high score for class $c_i$, and vice versa. 
\begin{algorithm}[ht!]
	\caption{ \textsc{The algorithm of our proposed XGNN.}}
	\label{al:1}
	\begin{algorithmic}[1]
		\State Given the trained GNNs for graph classification, denoted as $f(\cdot)$, we try to interpret it and set the target class as $c_i$. 
		\State Let $C$ define the candidate node set and $g(\cdot)$ mean our graph generator. We predefine the maximum generation step as $S_{max}$ and the number of Rollout as $m$. 
		\State Define the initial graph as $G_1$. 
		\For{step $t$ in $S_{max}$}
		\State Merge the current graph $G_t$ and the candidate set $C$.
		\State Obtain the action $a_t$ from the generator $g(\cdot)$ that $a_t=(a_{t,start}, a_{t,end})$ with Equation (\ref{eqn1_3}-\ref{eqn1_6}). 
		\State Obtain the new graph $G_{t+1}$ based on $a_t$.
		\State Evaluate $G_{t+1}$ with Equation (\ref{eq:r1}-\ref{eq:all}) and obtain $R_t$.
		\State Update the generator $g(\cdot)$ with Equation (\ref{eq:ce}).
		\If{$R_t < 0$}{
			roll back and set $G_{t+1}=G_{t}$. 
		}	
		\EndIf     
		\EndFor
	\end{algorithmic}
\end{algorithm}

In addition, the reward $R_{t,r}$ is obtained from graphs rules and is employed to encourage the generated graphs to be valid and human-intelligible. 
The first rule we employ is that only one edge is allowed to be added between any two nodes. Second, the generated graph cannot contain more nodes than the predefined maximum node number. In addition, we incorporate dataset-specific rules to guide the graph generation. For example, in a chemical dataset, each node represents an atom so that its degree cannot exceed the valency of the corresponding atom. When any of these rules is violated, a negative reward will be applied for $R_{t,r}$. Finally, by combining the $R_{t,f}$ and $R_{t,r}$, we can obtain the reward for step $t$ that 
\begin{equation}\label{eq:all}
R_{t}= R_{t,f} (G_{t+1})+ \lambda_1 \frac{\sum_{i=1}^{m}R_{t,f}(\mbox{Rollout}({G_{t+1}}))}{m} + \lambda_2 R_{t,r}, 
\end{equation}
where $\lambda_1$ and $\lambda_2$ are hyper-parameters. We illustrate the training procedure in Algorithm ~\ref{al:1}. Note that we roll back the graph $G_{t+1}$ to $G_{t}$ when the action $a_t$ is evaluated as not promising that $R_{t}<0$.

\section{Experimental Studies}

\subsection{Dataset and Experimental Setup}~\label{data}
We evaluate our proposed XGNN on both synthetic and real-world datasets. We report the summary statistics of
these datasets in Table~\ref{table:stat}. Since there is no existing work investigating model-level interpretations of GNNs, 
we have no baseline to compare with. Note that existing studies~\cite{ying2019gnn, baldassarre2019explainability} only focus on interpreting GNNs 
at example-level while ignoring the model-level explanations. Comparing with them is not expected since these example-level and model-level are two totally different interpretation directions.
\\
\textbf{Synthetic dataset}:  Since our XGNN generates model-level explanations for Deep GNNs, we build a synthetic dataset, known as Is\_Acyclic, where the ground truth explanations are available. The graphs are labeled based on if there is any cycle existing in the graph. The graphs are obtained using Networkx software package~\cite{hagberg2008exploring}.
The first class refers to cyclic graphs, including grid-like graphs, cycle graphs, wheel graphs, and circular ladder graphs. The second class denotes acyclic graphs, containing star-like graphs, binary tree graphs, path graphs and full rary tree graphs~\cite{storer2012introduction}. Note that all nodes in this dataset are unlabeled and we focus on investigating the ability of GNNs to capture graph structures.\\
\textbf{Real-world dataset}: We conduct experiments on the real-world dataset MUTAG. The MUTAG dataset contains graphs representing chemical compounds where nodes represent different atoms and edges represent chemical bonds. The graphs are labeled into two different classes according to their mutagenic effect on a bacterium~\cite{structure1991}. Each node is labeled based on its type of atom and there are seven possible atom types: Carbon, Nitrogen, Oxygen, Fluorine, Iodine, Chlorine, Bromine. Note that the edge labels are ignored for simplicity. For this dataset, we investigate the ability of GNNs to capture both graph structures and node labels.\\
\textbf{Graph classification models}: We train graph classification models using these datasets and then try to explain these models. These models share a similar pipeline that first learns node features using multiple layers of GCNs, then obtain graph level embeddings by averaging all node features, and finally employs fully-connected layers to perform graph classification. For the synthetic dataset Is\_Acyclic, we use the node degrees as the initial features for all nodes. Then we apply two layers of GCNs with output dimensions equal to 8, 16 respectively and perform global averaging to obtain the graph representations. Finally, we employ one fully-connected layer as the classifier. Meanwhile, for the real-world dataset MUTAG, since all nodes are labeled, we employ the corresponding one-hot representations as the initial node features. Then we employ three layers of GCNs with output dimensions equal to 32, 48, 64 respectively and average all node features. The final classifier contains two fully-connected layers in which the hidden dimension is set to 32. Note that for all GCN layers, we apply the GCN version shown in Equation (\ref{eq:1}). In addition, we employ Sigmoid as the non-linear function in GCNs for dataset Is\_Acyclic while we use Relu for dataset MUTAG.  These models are implemented using Pytorch~\cite{paszke2017automatic} and trained using Adam optimizer~\cite{kingma2014adam}. 
The training accuracies of these models are reported in Table~\ref{table:stat}, which show that the models we try to interpret are models with reasonable performance.\\
\begin{table}
	\centering \caption{Statistics and properties of datasets. Note that the edge number and node number are averaged numbers.} \label{table:stat}
	\begin{tabular}{  c | c | c | c | c}
		\hline
		Dataset & Classes  & \# of Edges & \# of Nodes & Accuracy\\ \hline\hline
		Is\_Acyclic  & 2 & 30.04  & 28.46 & 0.978\\ \hline
		MUTAG & 2 & 19.79  & 17.93 & 0.963 \\ \hline
		\hline
	\end{tabular}
\end{table}
\begin{figure*}[!ht]
	\centering
	\includegraphics[width=\textwidth]{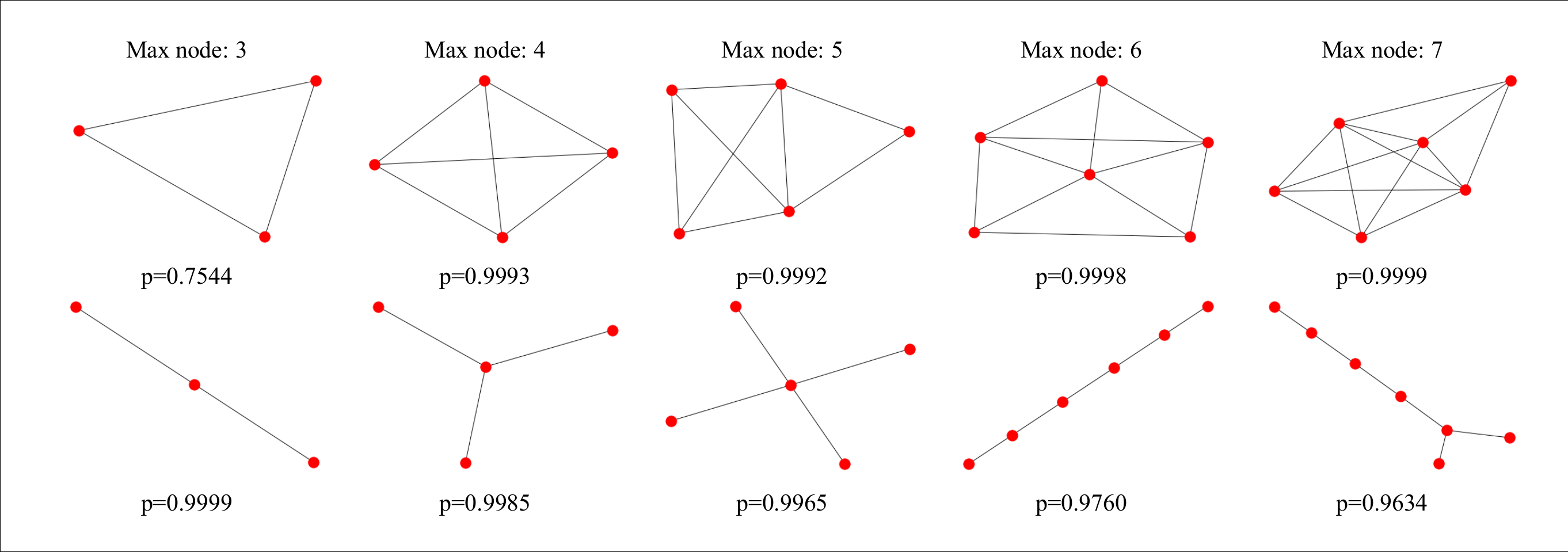}
	\caption{Experimental results for the synthetic dataset Is\_Acyclic. Each row shows our explanations for a certain class that the first row corresponds to the class cyclic while the second row explains the class acyclic. In each row, from left to right, we report the generated graphs with increasing maximum node number limits. In addition, we feed each generated graph to the pre-trained GCNs and report the predicted probability for the corresponding class.}
	\label{fig:res1}
\end{figure*}
\textbf{Graph generators}: For both datasets, our graph generators share the same structure. Our generator first employs a fully-connected layer to map node features to the dimension of 8. Then three layers of GCNs are employed with output dimensions equal to 16, 24, 32 respectively. The first MLPs 
consist of two fully-connected layers with the hidden dimension equal to 16 and a ReLU6 non-linear function. 
The second MLPs also have two fully-connected layers that the hidden dimension is set to 24 and ReLU6 is applied. 
The initial features for input graphs are the same as mentioned above.
For dataset Is\_Acyclic, we set $\lambda_1= 1$,  $\lambda_2= 1$, and $R_{t,r}= -1$ if the generated graph violates any graph rule. For dataset MUTAG, we set $\lambda_1= 1$,  $\lambda_2= 2$, and the total reward $R_{t}= -1$ if the generated graph violates any graph rule. In addition, we perform rollout $m=10$ times each step to obtain final graph rewards. The models are implemented using Pytorch~\cite{paszke2017automatic} and trained using Adam optimizer~\cite{kingma2014adam} with $\beta_1 =0.9$ and $\beta_2=0.999$. The learning rate for graph generator training is set to 0.01.

\subsection{Experimental Results on Synthetic Data}~\label{Synthetic}
We first conduct experiments on the synthetic dataset Is\_Acyclic where the ground truth is available. As shown in Table~\ref{table:stat}, the trained GNN classifier can reach a promising performance. 
Since the dataset is manually and synthetically built based on if the graph contains any circle, we can check if the trained GNN classifier makes predictions in such a way. We explain the model with our proposed XGNN and report the generated interpretations in Figure~\ref{fig:res1}. We show the explanations for the class ``cyclic'' in the first row and the results for the class ``acyclic'' in the second row. In addition, we also report different generated explanations by setting different maximum graph node limits. 

First, by comparing the graphs generated for different classes, we can easily conclude the difference that the explanations for the class ``cyclic'' always contain circles while the results for the class ``acyclic'' have no circle at all. Second, to verify whether our explanations can maximize the class probability for a certain class, as shown in Equation (\ref{form1}), we feed each generated graph to the trained GNN classifier and report the predicted probability for the corresponding class. The results show that our generated graph patterns can consistently yield high predicted probabilities. Note that even though the graph obtained for the class ``cyclic'' with maximum node number equal to 3 only leads to $p=0.7544$, it is still the highest probability for all possible graphs with 3 nodes. Finally, based on these results, we can understand what patterns can maximize the predicted probabilities for different classes. In our results, we know the trained GNN classifier very likely distinguishes different classes by detecting circular structures, which is consistent with our expectations. Hence, such explanations help understand and trust the model, and increase the trustworthiness of this model to be used as a circular graph detector. In addition, it is noteworthy that our generated graphs are easier to analyze compared with the graphs in the datasets. Our generated graphs have significantly fewer numbers of nodes and simpler structures, and yield higher predicted probabilities while the graphs from the dataset have an average of 28 nodes and 30 edges, as shown in Table~\ref{table:stat}.  

\subsection{Experimental Results on Real-World Data}~\label{real}
\begin{figure*}[!ht]
	\centering
	\includegraphics[width=\textwidth]{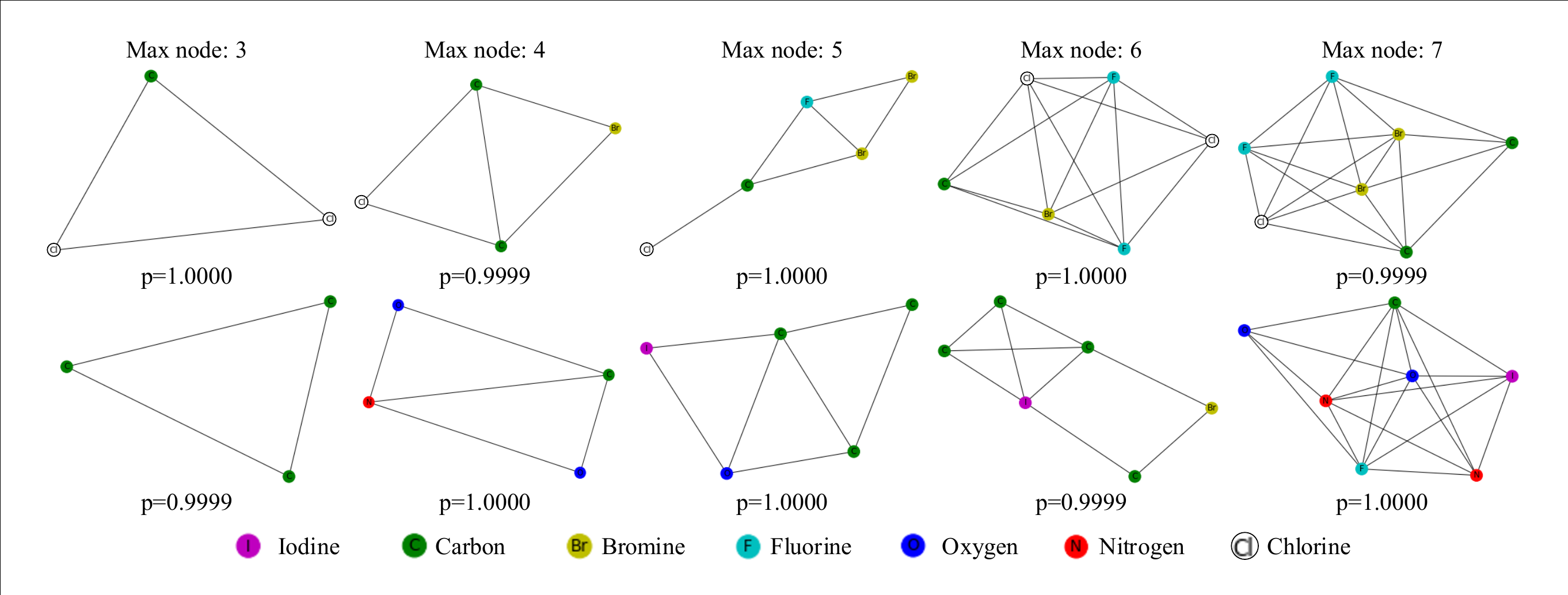}
	\caption{Experimental results for the MUTAG dataset. The first row reports the explanations for the class non-mutagenic while the second row shows results for the class mutagenic. Note that different node colors denote different types of atoms and the legend is shown at the bottom of the figure. All graphs are generated with the initial graph as a single Carbon atom.}
	\label{fig:res2}
\end{figure*}

\begin{figure*}[!ht]
	\centering
	\includegraphics[width=\textwidth]{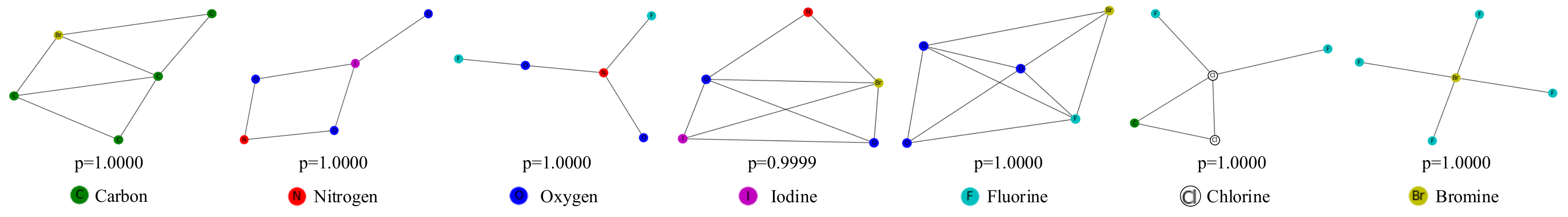}
	\caption{Experimental results for the MUTAG dataset. We fix the maximum node number limit as 5 and explore different initial graphs. Note that all graphs are generated for explaining the mutagenic class. For each generated graph, we show its predicted probability and corresponding initial graph at the bottom.}
	\label{fig:res3}
\end{figure*}

We also evaluate our proposed XGNN using real-world data. For dataset MUTAG, there is no ground truth for the interpretations. Since all nodes are labeled as different types of atoms, we investigate whether the trained GNN classifier can capture both graph structures and node labels.  We interpret the trained GNN with our proposed method and report selected results in Figure~\ref{fig:res2} and Figure~\ref{fig:res3}. Note that the generated graphs may not represent real chemical compounds because, for simplicity, we only incorporate a simple chemical rule that the degree of an atom cannot exceed its maximum chemical valency. In addition, since nodes are labeled, we can set the initial graphs as different types of atoms. 

We first set the initial graph as a single carbon atom and report the results in Figure~\ref{fig:res2}, since generally, any organic compound contains carbon~\cite{seager2013chemistry}. The first row reports explanations for the class ``non-mutagenic'' while the second row shows the results for the class ``mutagenic''. We report the generated graphs with different node limits and the GNN predicted probabilities. For the class ``mutagenic'', we can observe that carbon circles and $NO_2$ are some common patterns, and this is consistent with the chemical fact that carbon rings and  $NO_2$ chemical groups are mutagenic~\cite{structure1991}. 
Such observations indicate that the trained GNN classifier may capture these key graph patterns to make predictions. In addition, for the class ``non-mutagenic'', we observe the atom Chlorine is widely existing in the generated graphs and the combination of Chlorine, Bromine, and Fluorine always leads to ``non-mutagenic'' predictions. By analyzing such explanations, we can better understand the trained GNN model.

We also explore different initial graphs and report the results in Figure~\ref{fig:res3}. We fix the maximum node limit as 5 and generate explanations for the class ``mutagenic''. First, no matter how we set the initial graph, our proposed method can always find graph patterns maximizing the predicted probability of class ``mutagenic''. For the first 5 graphs, which means the initial graph is set to a single node of Carbon, Nitrogen, Oxygen, Iodine, or Fluorine, some generated graphs still have common patterns like carbon circle and $NO_2$ chemical groups. Our observations further confirm that these key patterns are captured by the trained GNNs. 
In addition, we notice that the generator can still produce graphs with Chlorine which are predicted as ``mutagenic'', which is contrary to our conclusion above. If all graphs with Chlorine should be identified as ``non-mutagenic'', such explanations show the limitations of trained GNNs. Then 
these generated explanations can provide guidance for improving the trained GNNs, for example, we may place more emphasis on the graphs Chlorine when training the GNNs. Furthermore, the generated explanations may also be used to retrain and improve the GNN models to correctly capture our desired patterns. 
Overall, the experimental results show that our proposed interpretation method XGNN can help verify, understand, and even help improve the trained GNN models.

\section{Conclusions}
Graphs neural networks are widely studied recently and have shown great performance for multiple graph tasks. However, graph models are still treated as black-boxes and hence cannot be fully trustable. It raises the need of investigating the interpretation techniques for graph neural networks. It is still a less explored area where existing methods only focus on example-level explanations for graph models. However, none of the existing work investigates the model-level interpretations of graph models which is more general and high-level. Hence, in this work, we propose a novel method, XGNN, to interpret graph models in the model-level. Specifically, we propose to find graph patterns that can maximize a certain prediction via graph generation. We formulate it as a reinforcement learning problem and generate graph pattern iteratively. We train a graph generator and for each step, it predicts how to add an edge into the current graph. In addition, we incorporate several graph rules to encourage the generated graphs to be valid and human-intelligible. Finally, we conduct experiments on both synthetic and real-world datasets to demonstrate the effectiveness of our proposed XGNN. Experimental results show that the generated graphs help discover what patterns will maximize a certain prediction of the trained GNNs. The generated explanations help verify and better understand if the trained GNNs make a prediction in our expected way. Furthermore, our results also show that the generated explanations can help improve the trained models.

\section*{ACKNOWLEDGMENTS}
This work was supported in part by National Science Foundation grants DBI-2028361, IIS-1714741, IIS-1715940, IIS-1845081, IIS-1900990 and Defense Advanced
Research Projects Agency grant N66001-17-2-4031.

\bibliographystyle{ACM-Reference-Format}
\bibliography{sample-base}

\end{document}